\documentclass[10pt,twocolumn,letterpaper]{article}

\usepackage{cvpr}              
\definecolor{cvprblue}{rgb}{0.21,0.49,0.74}
\usepackage[pagebackref,breaklinks,colorlinks,allcolors=cvprblue]{hyperref}

\def\paperID{14542} 
\def\confName{CVPR}
\def\confYear{2026}

\title{Sketch2CT: Multimodal Diffusion for Structure-Aware\\3D Medical Volume Generation}

\author{
Delin An and Chaoli Wang\\
University of Notre Dame\\
{\tt\small \{dan3, chaoli.wang\}@nd.edu}
}

\begin{document}
\maketitle

        
\begin{abstract}
Diffusion probabilistic models have demonstrated significant potential in generating high-quality, realistic medical images, providing a promising solution to the persistent challenge of data scarcity in the medical field. Nevertheless, producing 3D medical volumes with anatomically consistent structures under multimodal conditions remains a complex and unresolved problem. We introduce Sketch2CT, a multimodal diffusion framework for structure-aware 3D medical volume generation, jointly guided by a user-provided 2D sketch and a textual description that captures 3D geometric semantics. The framework initially generates 3D segmentation masks of the target organ from random noise, conditioned on both modalities. To effectively align and fuse these inputs, we propose two key modules that refine sketch features with localized textual cues and integrate global sketch-text representations. Built upon a capsule-attention backbone, these modules leverage the complementary strengths of sketches and text to produce anatomically accurate organ shapes. The synthesized segmentation masks subsequently guide a latent diffusion model for 3D CT volume synthesis, enabling realistic reconstruction of organ appearances that are consistent with user-defined sketches and descriptions. Extensive experiments on public CT datasets demonstrate that Sketch2CT achieves superior performance in generating multimodal medical volumes. Its controllable, low-cost generation pipeline enables principled, efficient augmentation of medical datasets. Code is available at \url{https://github.com/adlsn/Sketch2CT}.
\end{abstract}

\section{Introduction}
\label{sec:intro}


Medical image synthesis has become a pivotal research field aimed at overcoming the challenge of data scarcity in medical imaging. In contrast to natural images, collecting large-scale, high-quality, and well-annotated medical datasets is hindered by privacy restrictions, significant acquisition costs, and dependence on expert annotations. Meanwhile, deep learning models, which underpin recent breakthroughs in medical image analysis tasks such as 
segmentation~\cite{an2024sli2vol, an2025hierarchical, sciadvadw2825, Wolleb2022DiffusionSegmentation}, 
classification~\cite{Azizi2021BigSSL, Davila2024TransferLearning}, 
anomaly detection~\cite{Pinaya2022TransformerAnomaly, Wyatt2022AnoDDPM, Wolleb2022DiffusionAnomaly}, 
registration~\cite{Balakrishnan2019VoxelMorph, Siebert2025ConvexAdam}, and cross-modality translation~\cite{Lyu2022CTMRI}, 
require large volumes of labeled data for optimal performance. Thus, generating realistic and diverse synthetic medical images presents a promising avenue for enhancing model generalizability and robustness under limited data conditions.

Generative models, including generative adversarial networks (GANs)~\cite{Goodfellow2020GAN}
and denoising diffusion probabilistic models (DDPMs)~\cite{DDPM}, have achieved significant progress in image synthesis~\cite{Jiang2024FastDDPM}, inpainting~\cite{Ju2024BrushNet}, and super-resolution~\cite{Elsaid2019SRCNN}. Among these, diffusion models have garnered particular attention for their exceptional image quality, robust training, and strong capacity to represent complex data distributions. Conditional diffusion models~\cite{CDM} further enhance controllability by incorporating auxiliary modalities, such as text descriptions~\cite{Chen2018Text2Shape, Sanghi2022CLIPForge}, segmentation masks~\cite{KonzCDM24}, or sketches~\cite{Olsen2009SketchSurvey, Zhang2021Sketch2Model, Shen2021DeepSketchHair}, to guide generation across images, videos, and 3D point clouds. Building on their success in natural image domains, diffusion-based approaches have been adopted in medical imaging to help address the challenge of limited data.

Diffusion-based medical image synthesis can be categorized into 2D and 3D approaches, depending on the dimensionality of the generated data. 2D diffusion models~\cite{Fernandez2022SyntheticSeg} commonly produce individual image slices (e.g., CT or MRI), delivering visually realistic results but often failing to ensure anatomical continuity between adjacent slices. To improve realism, segmentation-guided 2D models~\cite{KonzCDM24} incorporate anatomical priors, enhancing local accuracy but still lacking global inter-slice consistency for full 3D volume reconstruction. Conversely, 3D diffusion models directly generate volumetric data to preserve spatial coherence; however, they require significant computational resources, limiting both achievable resolution and the scalability of the training process. Conditioning on 3D segmentation masks can enhance anatomical plausibility, although it introduces complexity and may destabilize the training process. Recently, latent diffusion models~\cite{Wang2022PretrainingI2I, ControlNet, PinayaTDCFNOC22} have emerged as a promising alternative by conducting the diffusion process in a compressed latent space, reducing computational costs while retaining high fidelity.

While unconditional diffusion models are valuable for general data augmentation, generating paired image-annotation data through conditional diffusion is essential for downstream medical image analysis tasks~\cite{MyronenkoYHX23, Marija2020, shiri2025ai, AortaDiff, VagenasGM23}. Segmentation-guided diffusion frameworks~\cite{KonzCDM24, Dorjsembe2024, AhnPCP25} have demonstrated strong results for medical image synthesis; however, they are constrained by their reliance on predefined segmentation masks, which limit both the diversity and controllability of the output. To address segmentation availability constraints, methods such as MedGen3D~\cite{HanXYKSYDX23} employ a two-stage pipeline in which random segmentation masks are generated before synthesizing corresponding medical volumes; however, the inherent randomness of these masks limits structural control.

To overcome these challenges, namely, (1) insufficient inter-slice consistency in 2D methods, (2) the high computational demands of full 3D diffusion, and (3) the limited controllability of current conditional models, we introduce Sketch2CT, a multimodal diffusion framework for structure-aware 3D medical volume generation. Sketch2CT comprises two stages: (1) a sketch-and-text conditioned latent diffusion model that produces anatomically consistent 3D segmentation masks from user-provided sketches and textual descriptions, and (2) a segmentation-conditioned latent diffusion model that synthesizes high-quality 3D CT volumes based on the generated masks. In our approach, sketches serve as an intuitive structural blueprint of the target anatomy, while textual descriptions provide complementary semantic and geometric details not captured by sketches alone. The fusion of these modalities enables the controllable, structure-preserving generation of realistic 3D medical volumes, as illustrated in Figure~\ref{fig:results_display}.

The principal contributions of this work are as follows:
\begin{itemize}
    \item 
    We effectively capture both local structural information from sketches and contextual semantics from text, thereby enhancing the model's attention to anatomically meaningful regions during segmentation generation.
    \item 
    We jointly fuse multimodal representations from local and global perspectives, thereby facilitating deep interaction between the sketch and text modalities and improving structural fidelity.
    \item 
    We enable low-cost, user-crafted sketches and text descriptions to generate high-quality, anatomically coherent 3D CT volumes, providing a practical, controllable solution for data augmentation.
\end{itemize}

\section{Related Work}
\label{sec:related}

\noindent\textbf{Generative models for natural image synthesis.}\ 
Generative models can be broadly categorized into two major branches: GANs and diffusion models. 
GANs learn a mapping from a latent noise distribution to the target data distribution through an adversarial game between a generator and a discriminator, and have achieved remarkable success in early image synthesis tasks. 
However, they often suffer from training instability and mode collapse, which limits the diversity and fidelity of the generated samples~\cite{Nichol2021ImprovedDDPM}. 
Diffusion models have recently emerged as a more stable and expressive alternative, formulating data generation as the reversal of a gradual noising process. 
They have demonstrated superior image quality, robustness in training, and diversity compared to GANs. 
In the natural image domain, diffusion models have achieved state-of-the-art performance in 2D image generation and have been extended to conditional generation by introducing auxiliary modalities such as text, segmentation maps, or sketches to control the synthesis process. 
Furthermore, several studies~\cite{Fan2017PointSetGen, Wang2018Pixel2Mesh, Guillard2021Sketch2Mesh, Pan2019DeepMesh} have explored generating 3D data from 2D or multimodal conditions, enabling geometry-aware synthesis and reconstruction. 
For instance, Wu et al.\ \cite{Wu2023SketchDiffusion} proposed a sketch- and text-conditioned diffusion framework to generate colored 3D point clouds, demonstrating the effectiveness of combining coarse structural priors with semantic information.

\noindent\textbf{Medical image synthesis with generative models.}\ 
Compared to the extensive research on natural image generation, diffusion-based medical image synthesis remains underexplored. 
Most existing studies~\cite{Abhishek2019Mask2Lesion, Baur2018MelanoGANs, Han2018BrainGAN, Sun2022HAGAN, You2023VarianceReduction} focus on 2D slice-level synthesis, where diffusion models generate large numbers of CT or MRI slices with considerable diversity. 
However, the lack of inter-slice consistency and anatomical plausibility limits their usability for data augmentation and downstream analysis. 
A common strategy to improve anatomical realism is to incorporate explicit anatomical priors, such as segmentation masks, to guide the synthesis process. 
Following this idea, Wang et al.\ \cite{Wang2022PretrainingI2I} employed pretrained latent diffusion models to generate medical images conditioned on segmentation maps, and Zhang et al.\ \cite{ControlNet} proposed ControlNet to enable conditional 2D image generation. 
Although these methods demonstrate the benefits of conditioning, they are primarily designed for natural images and do not adapt well to medical data. 
Konz et al.\ \cite{KonzCDM24} designed a segmentation-guided diffusion model (Seg-Diff) to synthesize medical images from segmentation masks, showing that the synthetic data can improve the performance of segmentation networks. 
Nevertheless, 2D-based approaches still struggle to maintain inter-slice continuity and cannot generate segmentation masks themselves.

To enhance controllability and data diversity, some studies generate segmentation masks as an intermediate step before image synthesis. 
For instance, Guibas et al.\ \cite{Guibas2017DualGAN} used a dual-GAN architecture to produce 2D segmentation masks for retinal image generation, and Fernandes et al.\ \cite{Fernandez2022SyntheticSeg} generated brain MRI images conditioned on masks from two coordinated generators. 
For volumetric synthesis, Subramaniam et al.\ \cite{Subramaniam2022TOFMRA} employed a Wasserstein GAN~\cite{Arjovsky2017WGAN} to jointly generate 3D patches and their corresponding masks, which improves spatial coherence. 
However, the high computational cost of full 3D diffusion models makes large-scale volume synthesis challenging, particularly when incorporating 3D segmentation masks as structural priors. 
Han et al.\ \cite{HanXYKSYDX23} addressed this issue with MedGen3D, a multi-conditional diffusion framework that generates both 3D segmentation masks and medical volumes. 
Their approach first synthesizes sub-volumes and subsequently infers the complete volume from the generated regions, partially mitigating computational overhead.

\noindent\textbf{Multimodal and structure-aware generation.}\ 
Existing medical image synthesis approaches largely focus on either 2D image generation or 3D volume synthesis conditioned on segmentation masks. 
With the success of conditional diffusion models in computer vision, recent research has begun to incorporate multimodal guidance, such as text, sketches, or multimodal imaging, to enhance controllability and interpretability. 
For text-based generation, Xu et al.\ \cite{Xu2018AttnGAN} proposed AttnGAN, and Zhang et al.\ \cite{Zhang2017StackGAN} introduced StackGAN to synthesize images from textual descriptions through multi-stage refinement. 
In 3D reconstruction, image-based conditioning is widely used to recover 3D shapes from single or multi-view images~\cite{Wang2021TransformerRecon}, such as Pix2Vox++~\cite{Xie2020Pix2Voxpp} and Pixel2Mesh~\cite{Wang2018Pixel2Mesh}. 
When image conditions are unavailable, sketches serve as an efficient and intuitive structural input. 
Guillard et al.\ \cite{Guillard2021Sketch2Mesh} utilized sketches to refine object shapes, and Zhang et al.\ \cite{Zhang2021Sketch2Model} reconstructed 3D models from single-view sketches. 
Recent works further demonstrate that combining sketches with text provides complementary information: sketches encode coarse geometry, while text conveys semantics and fine-grained details. 
For example, Wu et al.\ \cite{Wu2023SketchDiffusion} employed capsule attention~\cite{Sabour2017Capsule} to jointly process sketch and text features for controllable 3D point cloud generation.

Inspired by these advances, we aim to leverage multimodal information to guide the generation of 3D medical volumes. 
In particular, sketches offer intuitive structural guidance of the target anatomy, while text provides complementary semantic and morphological context. 
As diffusion models have proven superior to GANs in terms of image fidelity and stability, we adopt them as the backbone of our framework. 
We also employ a latent diffusion formulation that enables high-resolution 3D generation in a compressed latent space, thereby balancing efficiency and quality. 
Although latent compression may lose some fine-scale details, ensuring globally consistent and anatomically plausible 3D structure is more critical for medical image analysis tasks. 
To the best of our knowledge, Sketch2CT is the first framework to explore multimodal (sketch and text) diffusion for structure-aware 3D medical volume generation.

\section{Sketch2CT}

\begin{figure*}[htbp]
    \centering
    \includegraphics[width=0.875\linewidth]{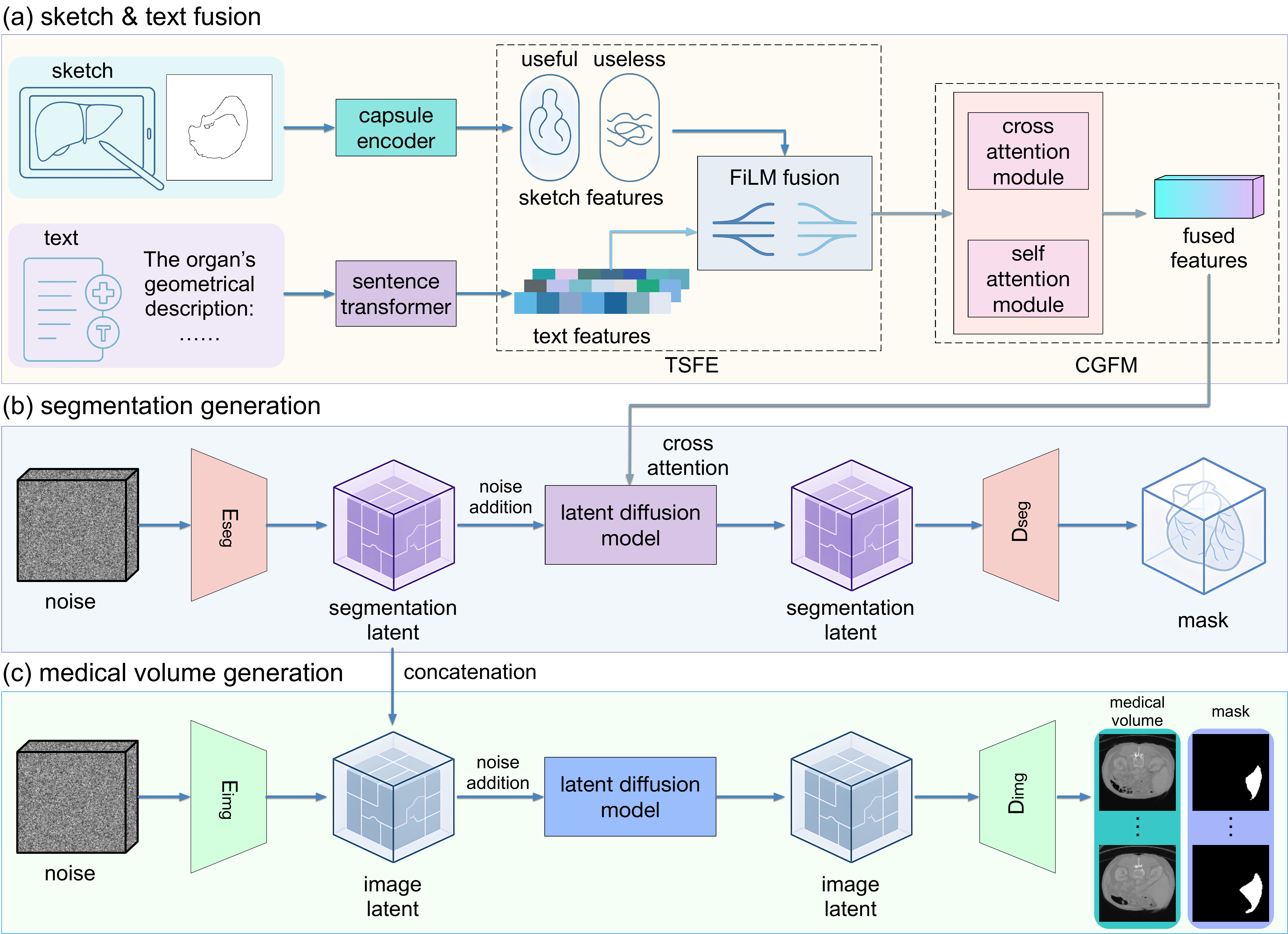}
    \caption{Overview of our \textit{Sketch2CT} framework.
        (a) A capsule-based sketch encoder and a sentence transformer extract structural and semantic features, which are fused via a FiLM module.
        (b) The fused representation conditions a segmentation latent diffusion model to generate 3D organ masks.
        (c) The predicted segmentation latent guides an image latent diffusion model to synthesize 3D medical volumes.}
    \label{fig:pipeline}
\end{figure*}

As shown in Figure~\ref{fig:pipeline}, our \textit{Sketch2CT} framework comprises two primary components: \textbf{segmentation mask generation} and \textbf{medical volume generation}. The first component is designed to synthesize 3D segmentation masks, conditioned on both sketches and textual descriptions. This stage includes four key modules: (1) \textbf{sketch extraction}, capturing the anatomical structure's outline; (2) \textbf{text acquisition}, providing complementary 3D semantic details; (3) \textbf{sketch-text feature fusion}, aligning structural and contextual information; and (4) \textbf{segmentation latent diffusion}, which generates a coherent 3D segmentation mask. The second component synthesizes realistic 3D medical volumes by conditioning a latent diffusion model on the segmentation representation. The detailed design and implementation of each module are described in the following subsections.

\subsection{Sketch Extraction}

To provide intuitive and structure-aware guidance for generation, we first extract 2D sketches from 3D anatomical segmentations. The objective is to obtain representative contour depictions that capture the characteristic shape and boundary geometry of each organ, while keeping the sketches lightweight, abstract, and easily interpretable for user interaction.
We begin by visualizing the 3D segmentation masks in a medical imaging environment (\textit{3D Slicer}~\cite{3DSlicer}) to obtain surface representations of the target anatomy. For volumetric organs such as liver and heart, 2D projections are captured along the \textbf{axial (transverse)} direction, which corresponds to the standard clinical viewing plane and effectively reveals the organ's overall morphology. For elongated, tubular structures such as the aorta, projections are instead taken along the \textbf{sagittal (longitudinal)} plane, which aligns with the vessel's principal axis and better preserves its continuous geometry. 

The rendered projections are subsequently transformed into sketches through a sequence of image-processing operations that emphasize structural boundaries while suppressing texture and shading. Specifically, we employ the \textit{PyVista}~\cite{Sullivan2019PyVista} library to render high-contrast surface projections with controlled lighting and orthogonal camera views, ensuring clear structural outlines. The rendered images are then processed using \textit{OpenCV} operations, including grayscale conversion, histogram equalization for contrast enhancement, bilateral filtering for edge-preserving smoothing, and Canny-based edge detection to extract salient contours. The resulting binary edge maps are refined using morphological operations to yield clean, continuous sketches that delineate the organ boundaries.
Through adjusting parameters such as edge detection thresholds and filtering kernel sizes, we can control the level of detail and abstraction in the sketches. 
These anatomy-aware sketches provide geometry-centric structural priors that condition the subsequent segmentation diffusion model.

\subsection{Text Acquisition}

Although sketches offer intuitive 2D structural information, they inherently lack depth and volumetric context, making it difficult to reconstruct a complete 3D segmentation mask from a single projection. Obtaining sketches from multiple views could mitigate this limitation, but would substantially increase user effort and compromise practical usability. To balance expressiveness with simplicity, we introduce text descriptions as a complementary modality to enhance the spatial context that sketches alone cannot provide.

To construct these textual representations, we first generate a series of 2D snapshots for each segmentation mask along the three principal anatomical axes: axial, coronal, and sagittal. These projections highlight different facets of the organ's morphology and serve as comprehensive visual references for its 3D structure. Simultaneously, we calculate a suite of physical and geometric metrics from the segmentation mask, such as volume, surface area, principal-axis lengths, maximum and minimum diameters, sphericity, compactness, and, when applicable, centerline length. Together, these quantitative attributes succinctly characterize the organ's overall shape, size, and spatial proportions. During training, textual descriptions are generated automatically from this visual and geometric information to serve as textual conditions. At inference time, however, users can freely edit or provide custom text descriptions, enabling flexible, interactive control over the generation process without requiring reference segmentations.

The snapshots and geometric metrics are provided together as prompts to a large language model (LLM) specialized in visual-language understanding (GPT-4o-mini, OpenAI)~\cite{Zhao2024LoRALand}. The model is directed to serve as an expert geometric describer, generating a structured textual summary of the organ's shape characteristics, including global form, surface smoothness, symmetry, and topological continuity, while explicitly excluding any diagnostic or clinical context. The generated output is a structured JSON description that retains only geometry-related information. This approach ensures the resulting text focuses solely on spatial and morphological features, treating the segmentation as a generic 3D object; thus, domain-specific medical LLMs are unnecessary.

Finally, the textual descriptions are transformed into high-dimensional embeddings via a pretrained sentence transformer~\cite{Reimers2019SentenceBERT}, capturing both global semantics and fine-grained geometric relationships conveyed in the text. These embeddings serve as textual conditioning features for the subsequent sketch-text fusion and segmentation-diffusion stages. By complementing sketch-based structural information with semantic representations of 3D geometry, the text modality enables more consistent, spatially aware mask generation.

\subsection{Sketch-Text Feature Fusion}

Although sketches provide an intuitive and compact representation of organ structure, their sparse, edge-only nature makes it challenging to extract stable, discriminative features for downstream 3D generation.
Unlike natural images, medical sketches contain limited texture and shading details, and the relevant anatomical boundaries are often discontinuous or faint due to projection or noise.
Directly encoding such sparse inputs with convolutional backbones can therefore lead to feature ambiguity and loss of structural cues.
To address these challenges, we design two complementary modules for multimodal fusion: a \textbf{text-enhanced sketch feature extractor (TSFE)} for local text-guided refinement, and a \textbf{cross-modal global fusion module (CGFM)} for global semantic alignment between sketch and text.

TSFE aims to enrich the sparse sketch representation with semantic priors derived from textual descriptions.
The sketch image is first encoded into a capsule-based embedding
$\mathbf{f}_s \in \mathbb{R}^{d_s}$ using a convolutional backbone followed by primary capsules and attention-based routing,
while the textual description is embedded as $\mathbf{f}_t \in \mathbb{R}^{d_t}$ using a pretrained sentence transformer.
To adaptively modulate the sketch embedding, we employ a feature-wise linear modulation (FiLM)~\cite{Perez2018FiLM} mechanism that generates scale and shift parameters from the text embedding
\begin{equation}
    \gamma, \beta = g(\mathbf{f}_t),
\end{equation}
where $g(\cdot)$ denotes a lightweight projection network.
The text-enhanced sketch feature is then obtained as
\begin{equation}
    \tilde{\mathbf{f}}_s = \gamma \odot \mathbf{f}_s + \beta,
\end{equation}
with $\odot$ indicating element-wise multiplication.
This text-guided modulation adaptively amplifies semantically relevant channels and suppresses irrelevant ones, producing a more informative and stable sketch embedding.

To achieve global semantic alignment and joint reasoning across modalities, we introduce a two-level attention mechanism for the CGFM, comprising text-guided cross-attention and sketch-guided self-attention.
Given the enhanced sketch features $\tilde{\mathbf{f}}_s$ and the text embedding $\mathbf{f}_t$, the cross-attention stage integrates fine-grained semantic features
\begin{equation}
    \mathbf{F}_{\text{local}} = \mathrm{Attention}(\tilde{\mathbf{f}}_s, \mathbf{f}_t, \mathbf{f}_t),
\end{equation}
capturing localized correspondences between sketch contours and textual semantics.
Subsequently, the self-attention stage aggregates these responses to form a global representation
\begin{equation}
    \mathbf{F}_{\text{global}} = \mathrm{SelfAttn}(\mathbf{F}_{\text{local}}),
\end{equation}
which summarizes the organ's overall geometry and semantics.
Finally, both representations are concatenated and projected to yield the joint multimodal feature
\begin{equation}
    \mathbf{z}_{\text{fusion}} = \mathrm{Proj}\big([\mathbf{F}_{\text{local}} \, \| \, \mathbf{F}_{\text{global}}]\big),
\end{equation}
which serves as the conditioning input for the segmentation latent diffusion model.

Together, TSFE enhances the sparse sketch representation through text-driven channel modulation, while CGFM aligns global semantics via hierarchical attention.
This two-stage fusion effectively bridges the modality gap, yielding a robust, geometry-aware multimodal embedding for generating anatomically consistent segmentations.

\subsection{Segmentation Latent Diffusion}
\label{subsec:MCSG}

Under the multimodal conditions provided by the fused sketch-text representation, we aim to generate anatomically coherent 3D segmentation masks using a latent diffusion framework.
Directly performing diffusion in voxel space is computationally demanding and memory-intensive for volumetric data.
To address this, we use the 3D AutoencoderKL function from \textit{MONAI} to project segmentation masks into a compact latent space, enabling efficient diffusion while preserving essential anatomical structures.

Formally, given a ground-truth segmentation volume $\mathbf{x}_0$, the autoencoder encodes it into a latent representation $\mathbf{z}_0 = E_{seg}(\mathbf{x}_0)$ and reconstructs it as $\hat{\mathbf{x}}_0 = D_{seg}(\mathbf{z}_0)$.
During training, the latent diffusion model learns a parameterized denoising process that progressively reconstructs $\mathbf{z}_0$ from Gaussian noise $\mathbf{z}_T \sim \mathcal{N}(0, \mathbf{I})$ across $T$ timesteps.

At each timestep $t$, the forward diffusion process adds Gaussian noise according to
\begin{equation}
    q(\mathbf{z}_t \mid \mathbf{z}_{t-1}) =
    \mathcal{N}\big(\sqrt{\alpha_t}\, \mathbf{z}_{t-1},\, (1 - \alpha_t)\mathbf{I}\big),
\end{equation}
where $\{\alpha_t\}_{t=1}^T$ defines a variance schedule controlling the noise intensity.
The reverse process is learned through a UNet-based denoising network $\epsilon_\theta$ conditioned on the multimodal feature $\mathbf{z}_{\text{fusion}}$
\begin{equation}
    p_\theta(\mathbf{z}_{t-1} \mid \mathbf{z}_t, \mathbf{z}_{\text{fusion}})
    = \mathcal{N}\big(\mu_\theta(\mathbf{z}_t, t, \mathbf{z}_{\text{fusion}}),
    \sigma_t^2 \mathbf{I}\big),
\end{equation}
where $\mu_\theta$ and $\sigma_t$ are learned mean and variance terms, and the conditioning is injected via cross-attention at multiple layers of the UNet.

Following the \textit{$v$-prediction} parameterization~\cite{Saharia2023}, the network is trained to predict the velocity term
\begin{equation}
    \mathbf{v}_t = \sqrt{\alpha_t}\, \boldsymbol{\epsilon} - \sqrt{1-\alpha_t}\, \mathbf{z}_0,
\end{equation}
where $\boldsymbol{\epsilon}$ denotes the Gaussian noise added at step $t$.
The training objective minimizes the mean squared error between the predicted and target velocities
\begin{equation}
    \mathcal{L}_{\text{diff}} =
    \mathbb{E}_{t,\,\mathbf{z}_0,\,\boldsymbol{\epsilon}}
    \big[\|\epsilon_\theta(\mathbf{z}_t, t, \mathbf{z}_{\text{fusion}}) - \mathbf{v}_t\|_2^2\big].
    \label{eqn:obj}
\end{equation}

During inference, random latent noise is iteratively denoised through the learned reverse process, guided by the sketch–text condition $\mathbf{z}_{\text{fusion}}$, to produce a clean segmentation latent $\hat{\mathbf{z}}_0$.
The pretrained autoencoder subsequently decodes this latent to yield the final 3D segmentation mask $\hat{\mathbf{x}}_0 = D(\hat{\mathbf{z}}_0)$.

By operating in the latent space and leveraging cross-modal conditioning, Sketch2CT efficiently synthesizes anatomically consistent segmentation masks that reflect both the structural constraints of the sketch and the semantic guidance of the text.

\subsection{Conditional Medical Volume Generation}

The final stage of Sketch2CT synthesizes 3D medical volumes conditioned on the generated segmentation structure.
While the segmentation diffusion model reconstructs anatomical geometry, this stage translates the structural representation into corresponding image appearances, thereby bridging the gap between geometry and texture.
To achieve this, we adopt a latent diffusion formulation similar to Section~\ref{subsec:MCSG} but extend it to incorporate the segmentation latent as a structural prior.

Specifically, the segmentation mask $\hat{\mathbf{x}}_{\text{seg}}$ obtained from the previous stage is first encoded into a compact latent representation $\mathbf{z}_{\text{seg}} = E_{\text{seg}}(\hat{\mathbf{x}}_{\text{seg}})$ using a pretrained 3D AutoencoderKL.
For the medical image volume $\mathbf{x}_{\text{img}}$, another AutoencoderKL is trained to map the medical volume into a perceptually meaningful latent space $\mathbf{z}_{\text{img}} = E_{\text{img}}(\mathbf{x}_{\text{img}})$.

During the diffusion process, the segmentation latent is concatenated with the noisy image latent at each timestep to guide the denoising trajectory
\begin{equation}
    \mathbf{z}_{t-1} = \epsilon_\theta\big(\mathbf{z}_t \, \| \, \mathbf{z}_{\text{seg}},\, t\big),
\end{equation}
where $\|$ denotes channel-wise concatenation.
This conditioning scheme ensures that the generated image remains anatomically aligned with the segmentation, while enabling the realistic synthesis of tissue texture.
The diffusion model is trained under the $v$-prediction parameterization following Equation~\ref{eqn:obj}. During inference, random noise is iteratively denoised under the segmentation guidance to produce a clean latent $\hat{\mathbf{z}}_{\text{img}}$, which is finally decoded into the voxel domain as $\hat{\mathbf{x}}_{\text{img}} = D_{\text{img}}(\hat{\mathbf{z}}_{\text{img}})$.

By conditioning the image diffusion model on the segmentation latent, Sketch2CT establishes a strong linkage between anatomical structure and visual realism. This design enables controllable, consistent generation of medical volumes, ensuring that synthesized images exhibit natural intensity variations while faithfully preserving the geometric fidelity of the generated segmentation.

\section{Results and Discussion}

We conduct our experiments on three public CT datasets and one MRI dataset:
(1) CHAOS liver (CT)~\cite{Kavur2021CHAOS}, which contains 20 CT scans annotated with liver;
(2) AVT aorta (CT)~\cite{Radl2022}, which contains 56 CT scans annotated with aorta;
(3) Decathlon liver (CT)~\cite{Antonelli2021MSD}, which contains 131 CT scans annotated with liver;
(4) Decathlon heart (MRI)~\cite{Antonelli2021MSD}, which contains 20 MRI scans annotated with heart.
For each dataset, we adopt an 8:2 train/test split.
All volumes are resampled to a unified resolution of $128{\times}128{\times}128$.

Our Sketch2CT consists of an autoencoder and latent diffusion models for both segmentation and medical volume generation.
The autoencoders use three resolution levels with channel widths $(32, 64, 64)$, a single residual block per level,
and spatial attention only in the deepest layer.
During diffusion training, the autoencoder is frozen and serves solely as the latent encoder-decoder.
The denoising backbone is a 3D UNet provided by \textit{MONAI} operating directly on the latent space.
The UNet employs channel widths $(32, 64, 64)$ with one residual block per level,
and applies spatial attention at the last two levels.
Cross-attention is enabled with a conditioning dimension of 1024, and one transformer layer is used at each attention-enabled level.
We train the model using a DDPM scheduler with 1000 steps and $v$-prediction parameterization.
The optimization uses Adam with a learning rate of $1\times10^{-4}$, a batch size of 10, and mixed-precision training.
The training runs for 300 epochs on an NVIDIA H200 GPU.

Additional experimental details, including the ablation studies of Sketch2CT, the effects of varying sketch granularity on mask generation, and other analyses, are provided in the supplementary materials.

\subsection{Synthetic Image Quality Evaluation}

To quantitatively assess the quality of our synthetic images, we compute two metrics: the Fréchet inception distance (FID) and the learned perceptual image patch similarity (LPIPS). Following established practices, such as those used in MedGen3D, both metrics are calculated based on multi-view projections of the generated volumes.
Table~\ref{tab:quantitative} compares Sketch2CT with three representative baselines: Med-DDPM~\cite{Dorjsembe2024}, MedGen3D~\cite{HanXYKSYDX23}, and Seg-Diff~\cite{KonzCDM24}.
All baselines are evaluated on a synthesized dataset constructed from two sources: (1) sketches and text extracted from ground-truth segmentations, and (2) manually created samples—low-cost hand-drawn sketches paired with lightly refined text descriptions. For each dataset, we produce 20 such hand-crafted sketch–text pairs.
Since the baselines cannot generate segmentation masks in a controlled manner, we use our synthesized masks as conditional inputs to ensure a fair comparison.

Sketch2CT achieves the best performance on most datasets.
On the challenging Decathlon heart dataset, where noise leads to higher FID and lower LPIPS for all methods, Sketch2CT still surpasses all baselines by a clear margin. For the CHAOS liver and AVT aorta datasets, it exhibits consistent gains in both fidelity and perceptual similarity.
Seg-Diff attains the best score on the Decathlon liver dataset, likely because its larger data volume favors a 2D diffusion model. However, its results often lack axial continuity. In contrast, Sketch2CT preserves full 3D spatial information while achieving image quality comparable to 2D methods.

\begin{table}[htbp]
    \centering
    \resizebox{\columnwidth}{!}{
        \begin{tabular}{lcc|cc|cc|cc}
            \hline
             & \multicolumn{2}{c|}{CHAOS liver (CT)}
             & \multicolumn{2}{c|}{AVT aorta (CT)}
             & \multicolumn{2}{c|}{Decathlon liver (CT)}
             & \multicolumn{2}{c}{Decathlon heart (MRI)}               \\
            \cline{2-9}
            method
             & FID$\downarrow$ & LPIPS$\uparrow$
             & FID$\downarrow$ & LPIPS$\uparrow$
             & FID$\downarrow$ & LPIPS$\uparrow$
             & FID$\downarrow$ & LPIPS$\uparrow$ \\
            \hline
            Med-DDPM
             & 114.4 & 0.220
             & 119.6 & 0.213
             & 115.3 & 0.207
             & 128.7 & 0.192                                     \\

            MedGen3D
             & 43.6  & 0.300
             & 47.1  & 0.294
             & 45.7  & 0.296
             & 96.8  & 0.248                                     \\

            Seg-Diff
             & 37.8  & 0.310
             & 38.9  & 0.313
             & \textbf{34.8} & \textbf{0.335}
             & 68.4  & 0.265                                     \\

            Sketch2CT
             & \textbf{33.7} & \textbf{0.332}
             & \textbf{36.9} & \textbf{0.321}
             & 36.5          & 0.328
             & \textbf{65.1} & \textbf{0.269}                     \\
            \hline
        \end{tabular}}
    \caption{Quantitative evaluation of synthetic images.
    We report FID (lower is better) and LPIPS (higher is better) across four datasets. 
    Since baseline models cannot generate conditional segmentations, we provide synthesized masks as their inputs.}
    \label{tab:quantitative}
\end{table}

\subsection{Benefits for Segmentation}

As described in \cite{KonzCDM24}, we assess the quality of generated medical images by measuring how well they preserve structural information relevant for segmentation. This evaluation criterion aligns well with our Sketch2CT framework, which explicitly generates 3D segmentation masks conditioned on sketch-text pairs. Then it synthesizes corresponding medical images, ensuring that anatomical structures are preserved throughout the process.

\begin{table*}[htbp]
    \centering
    \resizebox{2\columnwidth}{!}{
        \begin{tabular}{lcccccccc}
            \hline
             & \multicolumn{2}{c}{CHAOS liver (CT)}
             & \multicolumn{2}{c}{AVT aorta (CT)}
             & \multicolumn{2}{c}{Decathlon liver (CT)}
             & \multicolumn{2}{c}{Decathlon heart (MRI)}                                     \\
            \cline{2-9}
            model
             & Dice($m^{pred}_{gen}, m$)  & Dice($m^{pred}_{gen}, m^{pred}_{real}$)
             & Dice($m^{pred}_{gen}, m$)  & Dice($m^{pred}_{gen}, m^{pred}_{real}$)
             & Dice($m^{pred}_{gen}, m$)  & Dice($m^{pred}_{gen}, m^{pred}_{real}$)
             & Dice($m^{pred}_{gen}, m$)  & Dice($m^{pred}_{gen}, m^{pred}_{real}$)          \\
            \hline
            Med-DDPM
             & 0.501 & 0.492
             & 0.487 & 0.474
             & 0.512 & 0.495
             & 0.421 & 0.408                                                          \\

            MedGen3D
             & 0.814 & 0.797
             & 0.842 & 0.828
             & 0.821 & 0.803
             & 0.612 & 0.587                                                          \\

            Seg-Diff
             & 0.827 & 0.809
             & 0.866 & 0.861
             & 0.892 & 0.873
             & 0.638 & 0.601                                                          \\

            Sketch2CT
             & \textbf{0.868} & \textbf{0.852}
             & \textbf{0.894} & \textbf{0.887}
             & \textbf{0.912} & \textbf{0.904}
             & \textbf{0.642} & \textbf{0.614}                                        \\
            \hline
        \end{tabular}}
    \caption{Faithfulness of generated images to input masks. Dice is computed between segmentations predicted from generated images and the input masks (i.e., Dice($m^{pred}_{gen}, m$)),
        and between segmentations predicted from generated images and segmentations predicted from real images (i.e., Dice($m^{pred}_{gen}, m^{pred}_{real}$)).
        Sketch2CT consistently achieves the best fidelity across all datasets.}
    \label{tab:faithfulness}
\end{table*}

\noindent \textbf{Faithfulness to input masks.}\
Table~\ref{tab:faithfulness} reports the faithfulness of the generated images in relation to the input masks. For each method, we run an auxiliary segmentation network~\cite{KonzCDM24} (trained on real data) on the generated images and compute the Dice coefficient against (1) the input masks and (2) the segmentations predicted from the real images. Across all datasets, Sketch2CT achieves the highest Dice scores, demonstrating that the synthesized images most faithfully preserve spatial structure. 

\noindent \textbf{Downstream segmentation generalization.}\
To evaluate the effectiveness of synthetic data, we train a segmentation network using either (1) real training images or (2) synthetic images generated by various methods, and test the models on real datasets, as shown in Table~\ref{tab:downstream_segmentation}. Sketch2CT produces results closest to the real data baseline across all datasets, indicating that its generated images effectively replicate realistic anatomical details.

\begin{table}[htbp]
    \centering
    \resizebox{\columnwidth}{!}{
        \begin{tabular}{lcccc}
            \hline
             & CHAOS liver 
             & AVT aorta 
             & Decathlon liver 
             & Decathlon heart \\
             & (CT)
             & (CT)
             & (CT)
             & (MRI) \\
            \hline
            real training set
             & 0.897
             & 0.904
             & 0.912
             & 0.823                \\
            \hline
            Med-DDPM
             & 0.574
             & 0.561
             & 0.598
             & 0.412                \\

            MedGen3D
             & 0.814
             & 0.801
             & 0.823
             & 0.681                \\

            Seg-Diff
             & 0.826
             & 0.812
             & 0.887
             & 0.702                \\

            Sketch2CT
             & \textbf{0.893}
             & \textbf{0.889}
             & \textbf{0.904}
             & \textbf{0.711}       \\
            \hline
        \end{tabular}}
    \caption{Downstream segmentation performance (Dice) on real datasets.
        A segmentation model trained on synthetic images generated by Sketch2CT achieves the closest performance
        to the model trained on real data, demonstrating superior anatomical realism.}
    \label{tab:downstream_segmentation}
\end{table}

\begin{figure*}[htbp]
    \centering
    \includegraphics[width=1\linewidth]{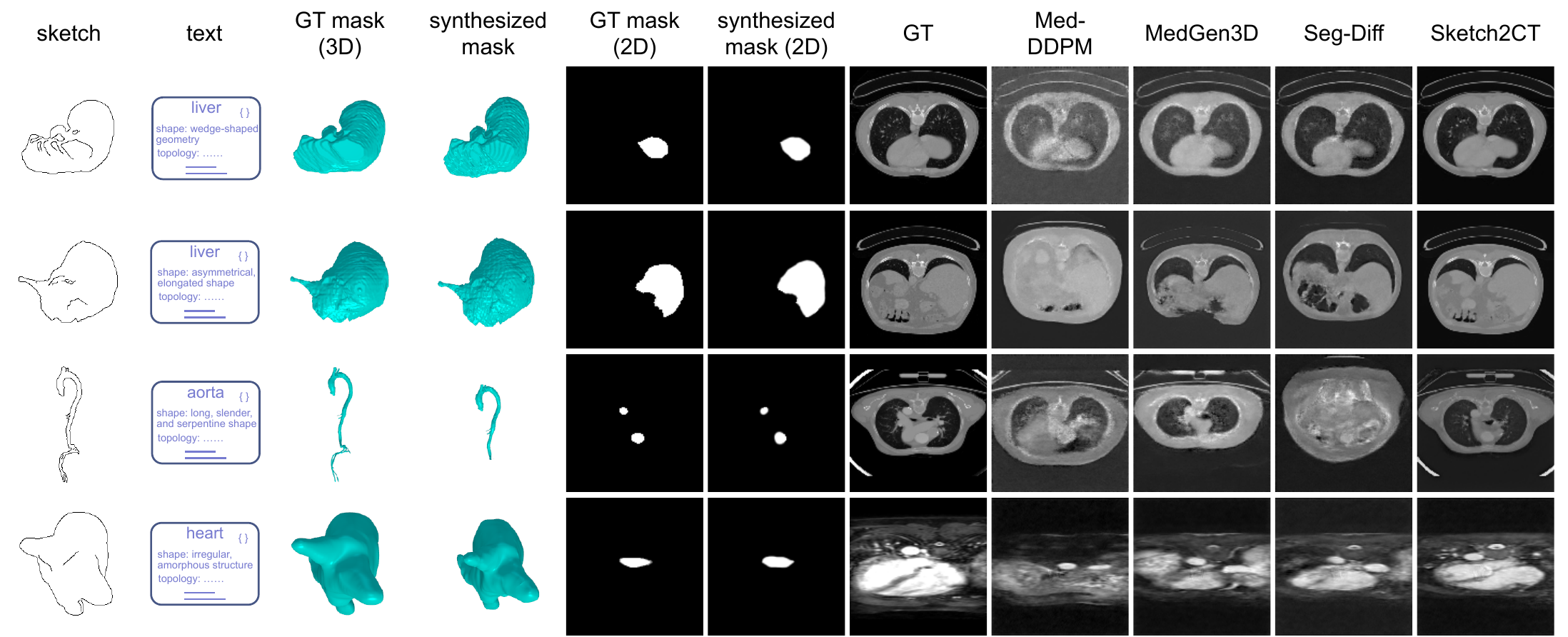}
    \caption{Qualitative comparison of baseline methods.
        For each case, we extract a sketch and text description from the ground-truth mask to encode organ geometry. These features guide the generation of 3D masks and the synthesis of 3D volumes.}
    \label{fig:baselines}
\end{figure*}

\subsection{Qualitative Results}

We also provide qualitative comparisons to complement the quantitative results and visually assess the anatomical plausibility of the synthesized segmentations and volumes. As shown in Figure~\ref{fig:baselines}, these examples reflect the trends observed in previous evaluations.
For each case, we extract a sketch and a text description from the ground-truth segmentation to encode the organs' geometry. We then generate segmentation conditioned on these features and use it to guide the generation of medical images. We compare the synthesized results from various methods. Med-DDPM and MedGen3D often lose structural details, while Seg-Diff shows promising results for the single slice. In contrast, Sketch2CT produces spatially coherent shapes with more precise boundaries.

\section{Conclusions and Future Work}

We have presented Sketch2CT, a multimodal diffusion framework that generates structure-aware 3D medical volumes from sketches and textual descriptions. The framework comprises a segmentation generator that reconstructs anatomical structures based on multimodal inputs and a volume generator that synthesizes realistic CT appearances guided by the segmentation latent. The proposed TSFE and CGFM modules effectively align sketch and text features with 3D anatomical semantics, thereby facilitating efficient, high-fidelity reconstruction.
Our experiments demonstrate that Sketch2CT enhances both image realism and data availability. A key advantage of this framework is its ability to utilize low-cost, user-created sketches and simple text descriptions to produce anatomically coherent segmentation masks and corresponding CT volumes. This level of control yields reliable synthetic data that enhances data augmentation and supports downstream segmentation tasks. However, the current implementation is limited to a few organs and focuses exclusively on the synthesis of single organs. In addition, two medical imaging experts qualitatively evaluated the generated results in terms of anatomical realism, structural continuity, and clinical plausibility, confirming their overall reliability. Future work will extend Sketch2CT to multi-organ generation, incorporate disease-specific sketch editing to simulate pathological variations, and conduct broader expert evaluations.

\noindent{\bf Acknowledgments.}\
This research was supported in part by the U.S.\ National Science Foundation through grants IIS-2101696, OAC-2104158, and IIS-2401144, and the U.S.\ National Institutes of Health through grant 7R01HL177814-02.

{
    \small
    \bibliographystyle{ieeenat_fullname}
    \bibliography{main}
}

\clearpage
\appendix
\setcounter{figure}{0}
\setcounter{table}{0}
\renewcommand{\thefigure}{A\arabic{figure}}
\renewcommand{\thetable}{A\arabic{table}}
\setcounter{page}{1}
\maketitlesupplementary

\section{Structured Semantic Text Description}

\begin{figure*}[htbp]
    \centering
    \includegraphics[width=0.9\linewidth]{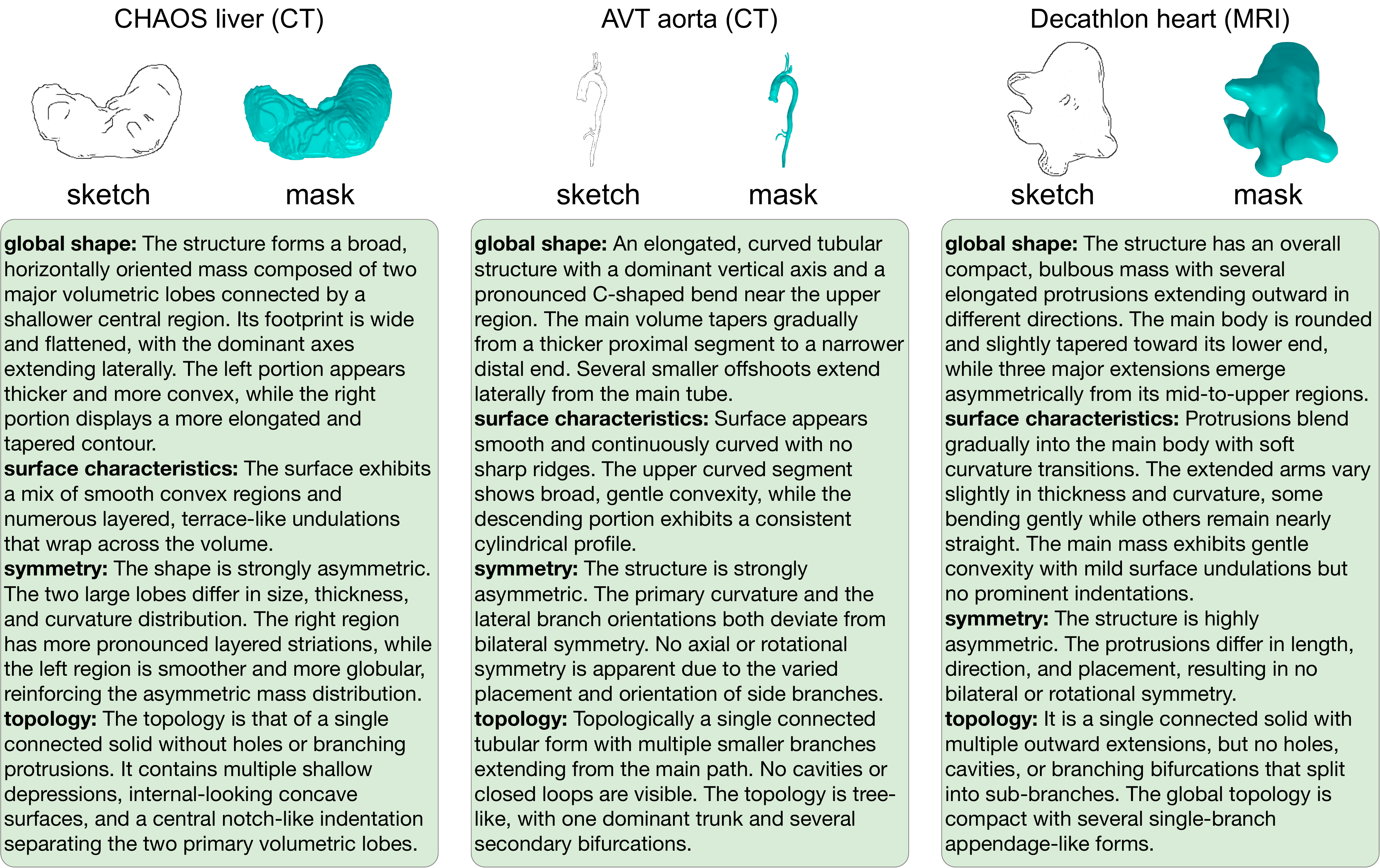}
    \caption{Examples of textual geometry descriptions used as the text-based conditioning input in Sketch2CT. For each organ, we show the input sketch, the generated mask, and the corresponding structured text description that captures the global shape, surface characteristics, symmetry, and topology.}
    \label{fig:text}
\end{figure*}

To obtain geometry-focused textual descriptions for multimodal conditioning, we provide GPT-4o-mini with (1) three canonical 3D renderings of each organ (axial, sagittal, and coronal views), and (2) a carefully designed prompt instructing the model to extract purely geometric information.
Examples of the text descriptions are presented in Figure~\ref{fig:text}.

\textbf{Prompt design.}\
GPT-4o-mini is directed to serve as an expert in the geometric interpretation of anatomical 3D structures. The prompt emphasizes strict geometric reasoning and forbids any clinical or physiological interpretation. In summary, the model is tasked with:

\begin{itemize}
    \item analyzing the organ's 3D geometry using the three provided views;
    \item describing the volumetric shape, dominant axes, convexity, and proportions;
    \item characterizing surface morphology, including curvature patterns, smoothness, ridges, protrusions, and indentations;
    \item assessing bilateral or rotational symmetry and quantifying asymmetry;
    \item identifying high-level topological traits such as cavities, branching, or major geometric landmarks; and
    \item providing additional structural features that could assist downstream geometry-aware diffusion models.
\end{itemize}

The prompt explicitly limits the model to geometric and morphological observations, instructing it to avoid using medical terminology, diagnoses, clinical relevance, or biological functions. The aim is to produce a text description that focuses solely on the structural characteristics of the input 3D shape.

\textbf{Output structure.}\
GPT-4o responds with a structured description organized into the following conceptual components:

\begin{itemize}
    \item an identifier for the input organ instance;
    \item a summary of global volumetric shape and coarse structural proportions;
    \item a characterization of surface-level geometric features;
    \item an analysis of symmetry or asymmetry patterns;
    \item a summary of key topological features; and
    \item high-level geometric guidance relevant for shape-conditioned generative modeling.
\end{itemize}

Although internally expressed in a structured format, we use only the distilled textual content for multimodal conditioning in the Sketch2CT framework.

\begin{figure*}[htbp]
    \centering
    \includegraphics[width=0.8\linewidth]{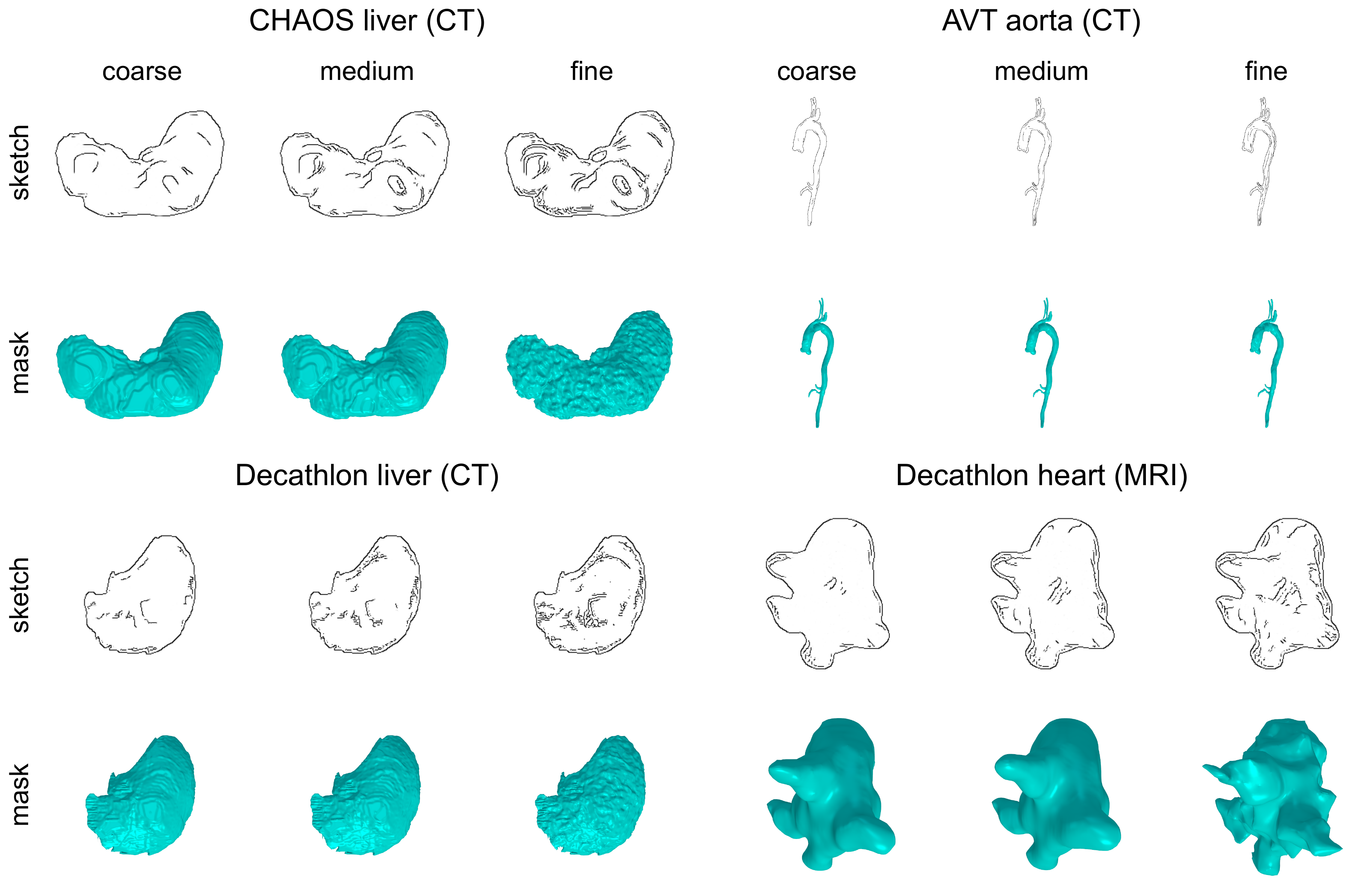}
    \caption{Effect of sketch granularity on segmentation mask generation. For each dataset, we vary the sketch detail in three levels, coarse, medium, and fine, and visualize the corresponding 3D masks produced by Sketch2CT. Increased sketch detail introduces more local structural variations, while the overall anatomical geometry remains consistent across granularity levels.}
    \label{fig:skechdetail}
\end{figure*}

\section{Sketch Granularity Analysis}

To investigate how sketch detail influences the quality of generated segmentation masks, we conduct a controlled analysis using sketches derived from 2D organ snapshots. As described in the main paper, sketch contours are generated using an edge-based extractor with a sensitivity parameter ranging from 0 to 10. Lower values produce sparse and coarse structural outlines, while higher values generate denser sketches with fine-grained details.

We select three representative parameter values across this range to create sketches with increasing levels of granularity. These sketches are then input into our segmentation generation module, while maintaining consistency with all other model components and conditions. The results are shown in Figure~\ref{fig:skechdetail}.
Our results reveal a clear trend:
\begin{itemize}
    \item \textbf{Coarse sketches} (low parameter settings) yield segmentation masks with smooth, global surfaces and minimal small-scale fluctuations. The generated structures faithfully capture the overall anatomical form, with stable topology and consistent volumetric shape.
    \item \textbf{Detailed sketches} (high parameter settings) introduce additional local variations, resulting in segmentation masks with slightly rougher or more irregular surface patterns. These high-detail sketches accurately capture fine contour variations and transfer them into the predicted mask.
\end{itemize}

Despite the surface-level differences, the global anatomical fidelity remains highly consistent across various levels of detail. The final reconstructed 3D medical volumes derived from these segmentation masks are visually similar, and quantitative metrics indicate minimal variation across different levels of sketch detail.
In particular, the Dice score between coarse and medium sketches fluctuates within $0.02\pm 0.01$. In contrast, the Dice difference between coarse and fine sketches remains similarly small at approximately $0.06\pm 0.02$, confirming that increased sketch detail leads to only marginal changes in segmentation quality.

Considering both performance and computational efficiency, we recommend using sketches with low to moderate granularity. Coarse sketches offer sufficient structural guidance for the multimodal diffusion model while minimizing unnecessary local noise and reducing preprocessing overhead. This supports a practical use case where users can provide simple, clean sketches without compromising the quality of downstream generation.


\section{Ablation Study}

To evaluate the contribution of the core components in Sketch2CT, we conduct an ablation study aligned with the modules defined in the main framework. Specifically, we examine the impact of removing: (1) TSFE, which refines sparse sketch embeddings via text-guided FiLM modulation; (2) CGFM, which performs global semantic alignment through hierarchical cross- and self-attention; and (3) the segmentation latent diffusion model, which reconstructs coherent 3D masks in latent space under multimodal conditioning. 
For each ablated variant, synthetic images are generated and used to train the same downstream segmentation network~\cite{Fernandez2022SyntheticSeg, ShinTTRSGM18} as in the main experiments. Table~\ref{tab:ablation} reports Dice scores on real test sets. Removing any single module leads to a consistent decrease in performance across all datasets, demonstrating that all three components play complementary roles in ensuring accurate multimodal alignment and anatomically faithful 3D mask generation. The full model achieves the best performance across all benchmarks.

\begin{table}[htbp]
    \centering
    \resizebox{\columnwidth}{!}{
        \begin{tabular}{lcccc}
            \hline
             & CHAOS liver & AVT aorta & Decathlon liver & Decathlon heart \\
             & (CT)        & (CT)      & (CT)            & (MRI)           \\
            \hline
            full model
             & \textbf{0.893}
             & \textbf{0.889}
             & \textbf{0.904}
             & \textbf{0.711}                 \\
            \hline
            w/o TSFE
             & 0.864
             & 0.859
             & 0.872
             & 0.683                          \\
            w/o CGFM
             & 0.825
             & 0.818
             & 0.831
             & 0.671                          \\
            w/o Seg-LDM
             & 0.642
             & 0.629
             & 0.633
             & 0.545                          \\
            \hline
        \end{tabular}}
    \caption{Ablation study of Sketch2CT. Removing TSFE, CGFM, or the segmentation latent diffusion model (Seg-LDM) reduces downstream segmentation performance, confirming that all components are essential for generating anatomically coherent and text-aligned 3D masks.}
    \label{tab:ablation}
\end{table}

\begin{figure}[htbp]
    \centering
    \includegraphics[width=\linewidth]{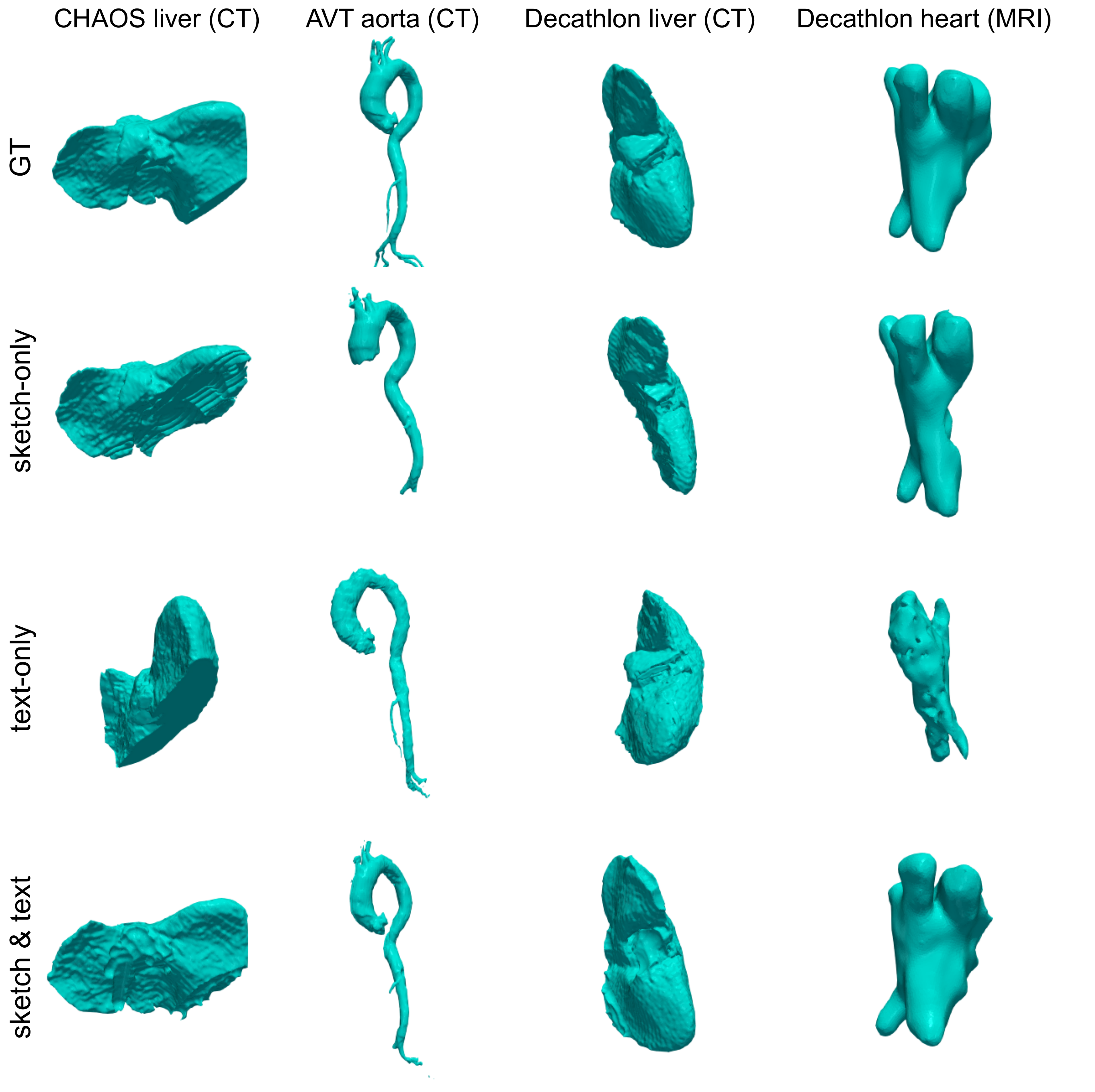}
    \caption{Comparison of segmentation masks generated using sketch-only and text-only conditions. Sketch-only guidance fails to recover full 3D geometry due to single-view ambiguity. In contrast, text-only guidance yields incorrect global shape and spatial placement, underscoring the need to combine sketches and text.}
    \label{fig:ablation}
\end{figure}

We further analyze the individual contributions of sketches and text by evaluating two additional variants: sketch-only and text-only conditioning. As shown in Figure~\ref{fig:ablation}, relying solely on sketches leads to incomplete or distorted 3D structures due to the inherent limitations of single-view contours, which lack depth information and volumetric context. Conversely, using text alone removes spatial constraints entirely, resulting in incorrect global shape, misplaced structures, and anatomically implausible geometries. These observations highlight that sketches and text provide complementary forms of guidance. Sketches anchor the spatial structure~\cite{Olsen2009SketchSurvey, Zhang2021Sketch2Model, Guillard2021Sketch2Mesh}, while text provides semantic and morphological context, both of which are essential for accurate and stable 3D mask generation.

\begin{figure*}[htbp]
    \centering
    \includegraphics[width=0.8\linewidth]{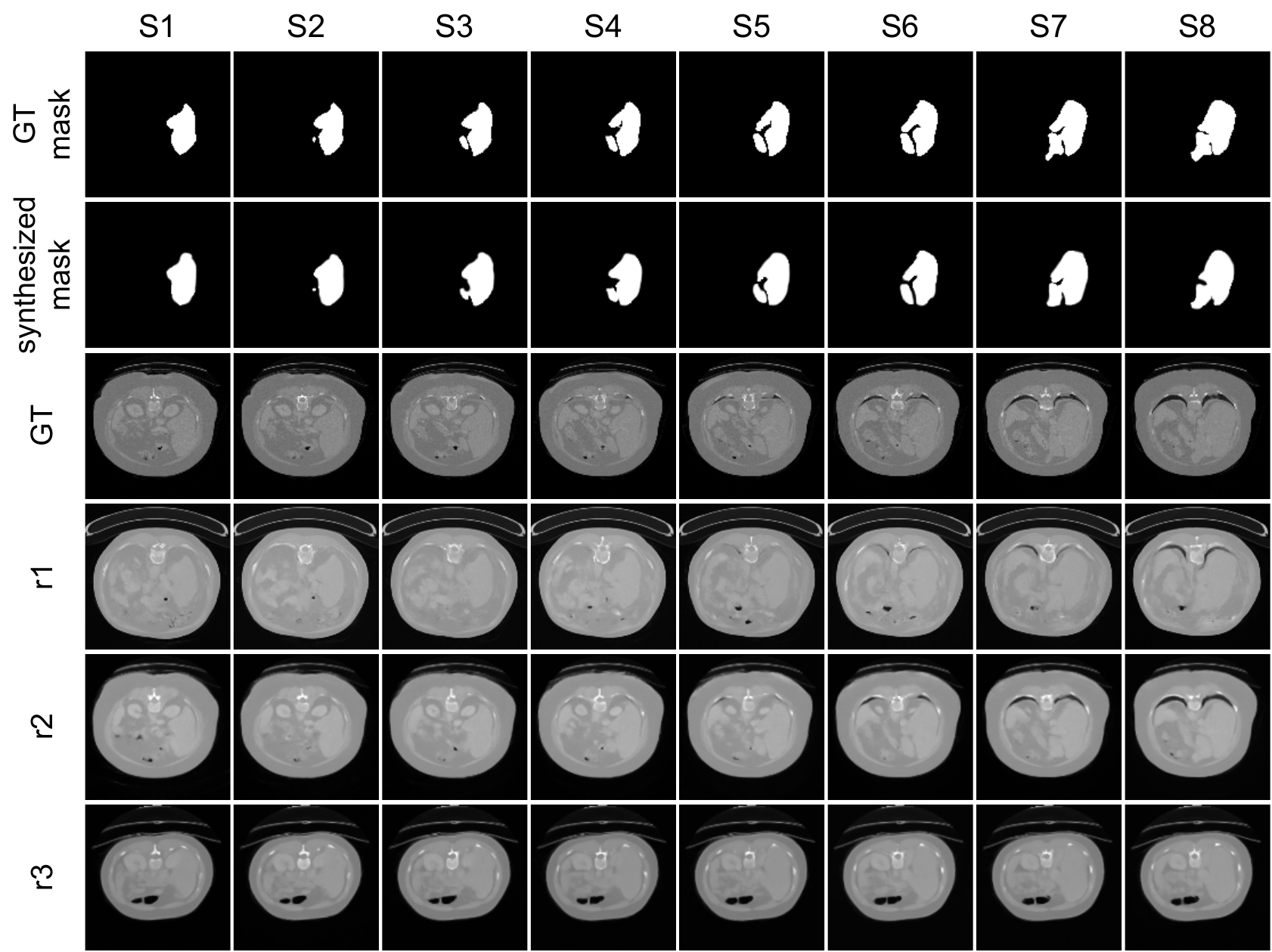}
    \caption{Qualitative diversity demonstration using a single Decathlon liver case. Under identical sketch and text conditions, three independent runs (r1-r3) generate anatomically consistent yet appearance-varying CT volumes, illustrating the stochastic diversity of Sketch2CT.}
    \label{fig:diversity}
\end{figure*}

\section{Qualitative Evaluation of Generative Diversity}

To qualitatively demonstrate the diversity of the synthesized volumes generated by Sketch2CT, we repeat the generation process three times under the same sketch and text conditions. As shown in Figure~\ref{fig:diversity}, each run produces anatomically consistent 3D segmentation masks that follow the shared multimodal conditioning, while the synthesized CT volumes exhibit natural variability in texture, intensity distribution, and fine-scale appearance. This stochasticity reflects the inherent randomness of the diffusion process, enabling Sketch2CT to produce diverse yet structurally faithful volumetric data. The qualitative results are consistent with the quantitative findings reported in the main paper, indicating that Sketch2CT maintains stable geometry while supporting realistic appearance-level variation, which is valuable for creating diverse synthetic datasets.

\end{document}